\documentclass[10pt,twocolumn,letterpaper]{article}

\usepackage{cvpr}
\usepackage{times}
\usepackage{epsfig}
\usepackage{graphicx}
\usepackage{amsmath}
\usepackage{amssymb}
\usepackage{booktabs,ragged2e}
\usepackage{subcaption}
\usepackage{graphicx}
\usepackage[nocompress,space]{cite}

\usepackage{authblk}

\usepackage{pifont}
\newcommand{\cmark}{\ding{51}}%
\newcommand{\xmark}{\ding{55}}%
\usepackage[flushleft]{threeparttable}

\usepackage[pagebackref=true,breaklinks=true,letterpaper=true,colorlinks,bookmarks=false]{hyperref}

\cvprfinalcopy 

\def\cvprPaperID{****} 

\begin{document}

\title{DALES: A Large-scale Aerial LiDAR Data Set for Semantic Segmentation}

\author[]{Nina Varney}
\author[]{Vijayan K. Asari}
\author[]{Quinn Graehling}
\affil[]{Department of Electrical Engineering, University of Dayton\\
{\tt\small \{varneyn1, vasari1, graehlingq1\}@udayton.edu}}
\renewcommand\Authands{ and }

\maketitle

\begin{abstract}
We present the Dayton Annotated LiDAR Earth Scan (DALES) data set, a new large-scale aerial LiDAR data set with over a half-billion hand-labeled points spanning 10 $km^2$ of area and eight object categories. Large annotated point cloud data sets have become the standard for evaluating deep learning methods. However, most of the existing data sets focus on data collected from a mobile or terrestrial scanner with few focusing on aerial data. Point cloud data collected from an Aerial Laser Scanner (ALS) presents a new set of challenges and applications in areas such as 3D urban modeling and large-scale surveillance. DALES is the most extensive publicly available ALS data set with over 400 times the number of points and six times the resolution of other currently available annotated aerial point cloud data sets.  This data set gives a critical number of expert verified hand-labeled points for the evaluation of new 3D deep learning algorithms, helping to expand the focus of current algorithms to aerial data. We describe the nature of our data, annotation workflow, and provide a benchmark of current state-of-the-art algorithm performance on the DALES data set.
\end{abstract}

\section{Introduction}

\begin{figure}[t]
\begin{center}
\includegraphics[width=0.9\linewidth, trim={0 2cm 0 8.5cm},clip]{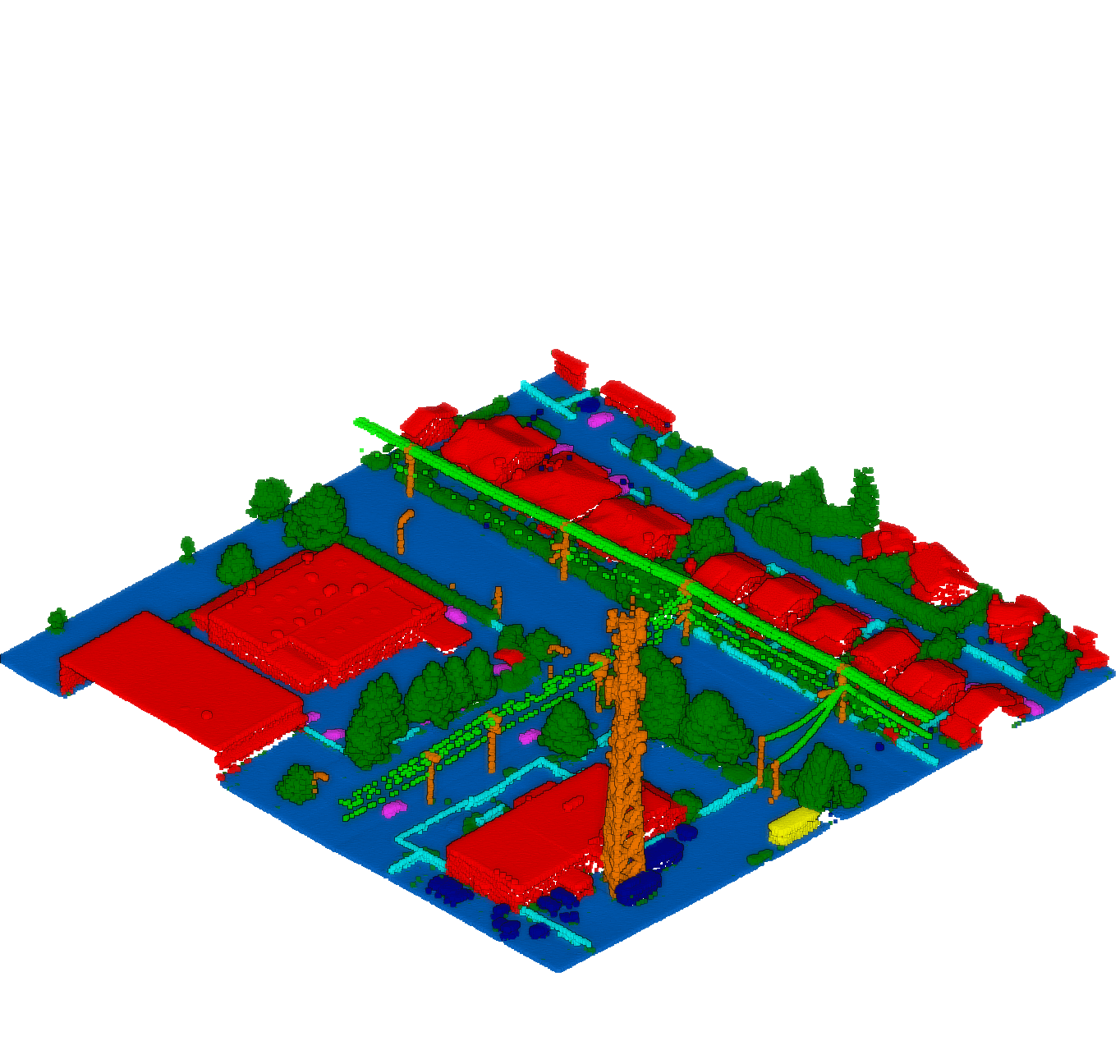}
\includegraphics[width=0.9\linewidth, trim={0 0 0 1cm},clip]{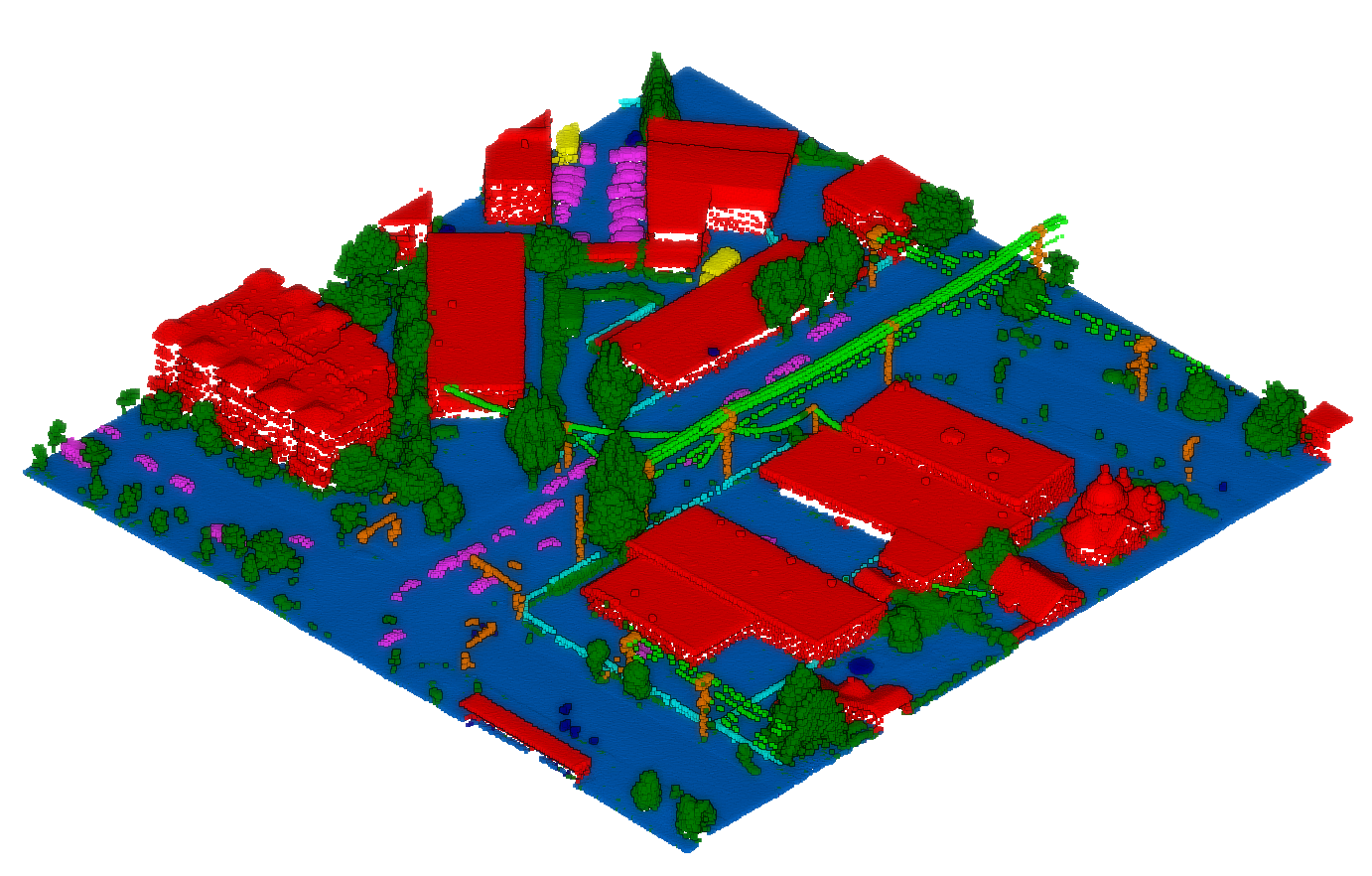}
\caption{Random sections taken from the tiles of our DALES data set. Each point is labeled by object category; ground (blue), vegetation (dark green), power lines (light green), poles (orange), buildings (red), fences (light blue), trucks (yellow), cars (pink), unknown (dark blue)}
\label{fig:1}
\end{center}
\end{figure}
The recent surge in autonomous driving and its use of LiDAR as part of the sensor suite has resulted in an enormous boost in research involving 3D data. Previously, the cost of LiDAR was a barrier to its widespread usage, but this barrier has been removed with the introduction of low-cost LiDAR. As the price of these devices continues to drop, there will be more research into developing algorithms to process LiDAR data, with the most prominent task being the semantic segmentation of LiDAR scenes. Semantic segmentation is a low-level product required for a variety of applications, including wireless signal mapping, forest fire management, terrain modeling, and public utility asset management.

Conducting deep learning on point clouds adds additional layers of complexity. The additional dimensionality greatly increases the number of parameters in the network. At the same time, the unstructured nature of the point cloud makes it incompatible with the convolutions deployed in traditional 2D imagery. PointNet++ and its predecessor PointNet \cite{pointnet++,pointnet} attempt to rectify this issue by mapping the 3D points into a higher dimensional feature space. This structured feature space allows the network to operate directly on the 3D points themselves instead of transforming them into an intermediate representation, such as voxels. Since the debut of PointNet, there has been a huge advancement and interest in the development of deep learning networks to work with unordered points.

The increased interest in working with LiDAR data, as well as advancement in methodology to operate directly on 3D points, has created a need for more point cloud data sets. There has been a wide range of data sets released for semantic segmentation from ground-based laser scanners, either static Terrestrial Laser Scanners (TLS) or Mobile Laser Scanners (MLS). These ground-based data sets include the Paris-Lille-3D, Sydney Urban Objects, and Semantic 3D data sets \cite{paris-lille, sydney,semantic3d}. However, there is a significant gap when considering the available aerial LiDAR data sets. For comparison, the ISPRS 3D Semantic Labeling data set \cite{isprs} is the primary available airborne data set with just over one million points.

There are several critical differences between aerial laser scanners (ALS) and other types of laser scanners. The first is the sensor orientation: ground-based laser scanners have a lateral sensor orientation, while aerial laser scanners collect data from a nadir orientation. This difference in sensor position means that, for the same scene, the resulting point clouds have significant differences both in point location and areas of occlusion. For example, buildings can be represented by the facade or the roof, depending on the orientation of the sensor. Another key difference is the point resolution: ground-based scanners, especially static scanners, have a much higher resolution but tend to lose resolution at areas further away from the scanner.

In comparison, the resolution in aerial LiDAR is lower overall but more consistent across a large area. The applications and object types also differ between ground-based and aerial sensors. Ground-based sensors have relatively small scenes, with a maximum range of around 200 meters. In contrast, airborne sensors can collect broad swaths of data over miles, making it appropriate for surveying, urban planning, and other Geographic Information System (GIS) applications. Finally, aerial LiDAR can be significantly more expensive due to the cost of both the sensor and the collection.

The above differences stress the importance of developing an ALS data set similar to those already existing for ground-based sensors. Inspired by the initial work of the ISPRS 3D Semantic Labeling data set, our goal is to present a critical number of labeled points for use within the context of a deep learning algorithm. Our objective is to expand the focus of current semantic segmentation algorithm development to include aerial point cloud data.

We present the Dayton Annotated LiDAR Earth Scan (DALES) data set, a semantic segmentation data set for aerial LiDAR. DALES contains forty scenes of dense, labeled aerial data spanning eight categories and multiple scene types including urban, suburban, rural, and commercial. This data set is the largest and densest publicly available semantic segmentation data set for aerial LiDAR. We focus on object categories that span a variety of applications. We work to identify distinct and meaningful object categories that are specific to the applications, avoiding redundant classes, like high and low vegetation. We have meticulously hand labeled each of these point clouds into the following categories: ground, vegetation, cars, trucks, poles, power lines, fences, buildings, and unknown. Figure \ref{fig:1} shows examples of object labeling. We split the data set into roughly 70\% training and 30\% testing. For ease of implementation, we provide this data set in a variety of input formats, including .las, .txt, and .ply files, all structured to match the input formats of other prominent data sets.

In addition to providing the data, we evaluated six state-of-the-art algorithms: KPConv \cite{kpconv}, ConvPoint \cite{convpoint}, PointNet++ \cite{pointnet++}, PointCNN \cite{pointcnn}, ShellNet \cite{shellnet} and Superpoint Graphs \cite{superpoint}. We evaluated these algorithms based on mean Intersection Over Union (IoU), per class IoU, Class Consistency Index (CCI), and overall accuracy. We also make recommendations for evaluation metrics to use with point cloud data sets. Our contributions are as follows:
\begin{itemize}
\itemsep-0.1cm
\item Introduce the largest publicly available, expert labeled aerial LiDAR data set for semantic segmentation
\item Test state-of-the-art deep learning algorithms on our benchmark data set
\item Suggest additional evaluation metrics to measure algorithm performance
\end{itemize}

\section{Related Work}

\begin{table*}[htbp]
\begin{center}
\caption{Comparison of 3D Data Sets}
\label{tab:2}
\setlength\tabcolsep{6pt} 
\begin{tabular}{c c c c c c c}
\toprule
     Name& Sensor Platform & RGB & Number of Points & Number of Classes\\
\midrule
     DALES (Ours)& aerial & \xmark & 505M &8\\
     ISPRS \cite{isprs}&aerial & \xmark & 1.2M &9\\
     \hline
     Sydney Urban Objects \cite{sydney}&mobile &\xmark & 2.3M & 26\\
     IQmulus \cite{iqmulus} &mobile&\xmark & 300M&22\\
     Oakland \cite{oakland}&mobile & \xmark&1.6M&44\\
     Paris-rue-Madame \cite{parisrue}& mobile &\xmark &20M&17\\
     Paris-Lille-3D \cite{paris-lille}&mobile & \xmark & 143M&50\\
     \hline
     Semantic 3D \cite{semantic3d}& static &\cmark &4000M&8\\
\bottomrule
\end{tabular}
\label{tab:comparison}
\smallskip
\scriptsize
\end{center}
\end{table*}
The development of any supervised deep learning method requires a high quality labeled data set. Traditionally, the field has relied on benchmark data sets to judge the success of a network and compare performance across different methods. Large-scale benchmark data sets, such as ImageNet \cite{imagenet} and COCO \cite{coco}, have become the standard for 2D imagery, containing pre-separated training and testing data to evaluate performance between methods. This has expanded into other data types, including point clouds \cite{nyu1,nyu2,s3dis,sunrgb,scannet,modelnet,synthcity, Matterport3D,kitti}.

LiDAR provides rich spatial data with a higher level of accuracy than other sensors, like RGB-D. Also, the cost of LiDAR sensors has decreased dramatically in recent years. As time goes on, more architectures have switched to using LiDAR data for training and testing. This switch to LiDAR data has meant a drastic increase in the demand for large-scale data sets captured with laser scanners. Table \ref{tab:comparison} presents a non-exhaustive list of some of the most popular laser scanning data sets. Most of these data sets, such as Sydney Urban Objects, Paris-rue-Madame, and Paris-Lille-3D \cite{sydney,parisrue, paris-lille, oakland} provide point cloud data from a sensor mounted on a moving vehicle traveling through urban environments. These MLS data sets have a much lower point density than a static sensor.

One publicly available 3D data set, Semantic 3D \cite{semantic3d}, contains outdoor scene captures with a notably high density of points. The scenes, collected with a TLS or static scanner, depict several different urban environments ranging from churches to small buildings, reaching a total of 4 billion points.  Semantic 3D also provides eight labeled classes: man-made terrain, natural terrain, high vegetation, low vegetation, buildings, hardscape, scanning artifacts, and cars. This data set was groundbreaking and provided one of the first significant benchmarks in high-density outdoor 3D data sets. Although dense data sets exist for ground and mobile sensors, they cannot be used for aerial LiDAR applications because of fundamental differences in the sensor orientation, resolution consistency, and areas of occlusion. Research in the field of aerial LiDAR requires a new type of annotated data, specific to aerial sensors.

The ISPRS 3D Semantic Labeling data set \cite{isprs} provides one of the only annotated data sets collected from an ALS sensor. These scenes depict point cloud representations of Vaihingen, Germany, consisting of three views with nine labeled point categories: power line, low vegetation, impervious surfaces, car, fence/hedge, roof, facade, shrub, and tree. The point clouds for each of these captures has a point density of 5-7 ppm and around 1.2 million points total. Though this data set was notable in providing one of the first sets of high quality labeled point cloud data captured from an aerial view, both the resolution and number of points are insufficient for deep learning applications.
\section{DALES: The Data Set}

We offer a semantic segmentation benchmark made exclusively of aerial LiDAR data. For each point within an unordered point set, a class label can be inferred. We review the data collection, point density, data preparation, preprocessing, 3D annotation, and final data format.

\subsection{Initial Data Collection}
The data was collected using a Riegl Q1560 dual-channel system flown in a Piper PA31 Panther Navajo. The entire aerial LiDAR collection spanned 330 $km^2$ over the City of Surrey in British Columbia, Canada and was collected over two days. The altitude was 1300 meters, with a 400$\%$ minimum overlap. The final data projection is UTM zone 10N, the horizontal datum is NAD83, and the vertical datum is CGVD28, using the metro Vancouver Geoid. Trimble R10 GNSS receivers collected ground control points with an accuracy within 1-2 cm.

Each area collected a minimum of 5 laser pulses per meter in the north, south, east, and west direction, providing a minimum of 20 ppm and minimizing occlusions from each direction. An accuracy assessment was performed using the ground control points along with a visual inspection, matching the corners and hard surfaces from each pass. The mean error was determined to be $\pm$ 8.5 cm at 95\% confidence for the hard surface vertical accuracy.

Along with the projected LiDAR information, a Digital Elevation Model (DEM) or bare earth model was calculated by a progressive triangulated irregular network (TIN) interpolation using Terrascan software. The DEM was then manually checked for inconsistencies and also cross-referenced with ground control points from the original collection. The final DEM resolution is 1 meter.

\subsection{Point Density}
One of the main differences between a static sensor and an aerial sensor is the existence of multiple returns. The distance between the object and the sensor causes the diameter of the pulse to be much larger than it would be in a ground or mobile-based sensor. The large diameter may cause the pulse to hit several different objects at distinct distances from the sensor. For this reason, a single laser pulse can result in multiple points in the final point cloud, allowing these sensors to achieve greater resolution, especially in high-frequency areas such as vegetation. In this data collection, we track as many as four returns from one single pulse. While this phenomenon increases spatial resolution during collection, it also introduces a unique difference from ground-based LiDAR and highlights a need for additional data sets to improve deep learning models. In this particular data set, the minimum guaranteed resolution during collection was 20 ppm for first returns (i.e., one point per pulse). We measured an average of 50 ppm, after initial noise filtering, when including all returns.

\begin{figure*}[htbp]
\centering
\begin{subfigure}{.33\textwidth}
  \centering
  \includegraphics[width=1\linewidth]{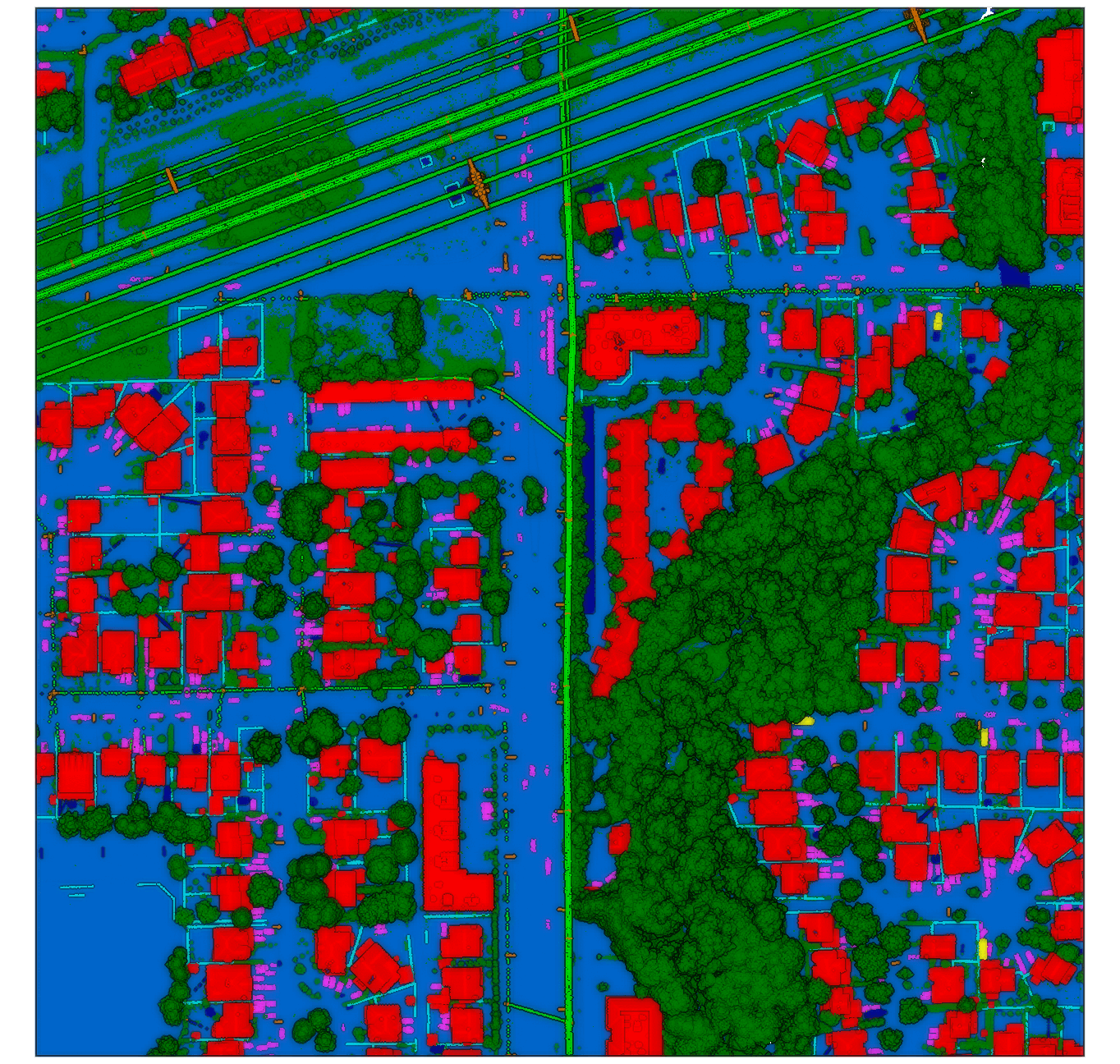}
\end{subfigure}%
\begin{subfigure}{.33\textwidth}
  \centering
  \includegraphics[width=1\linewidth]{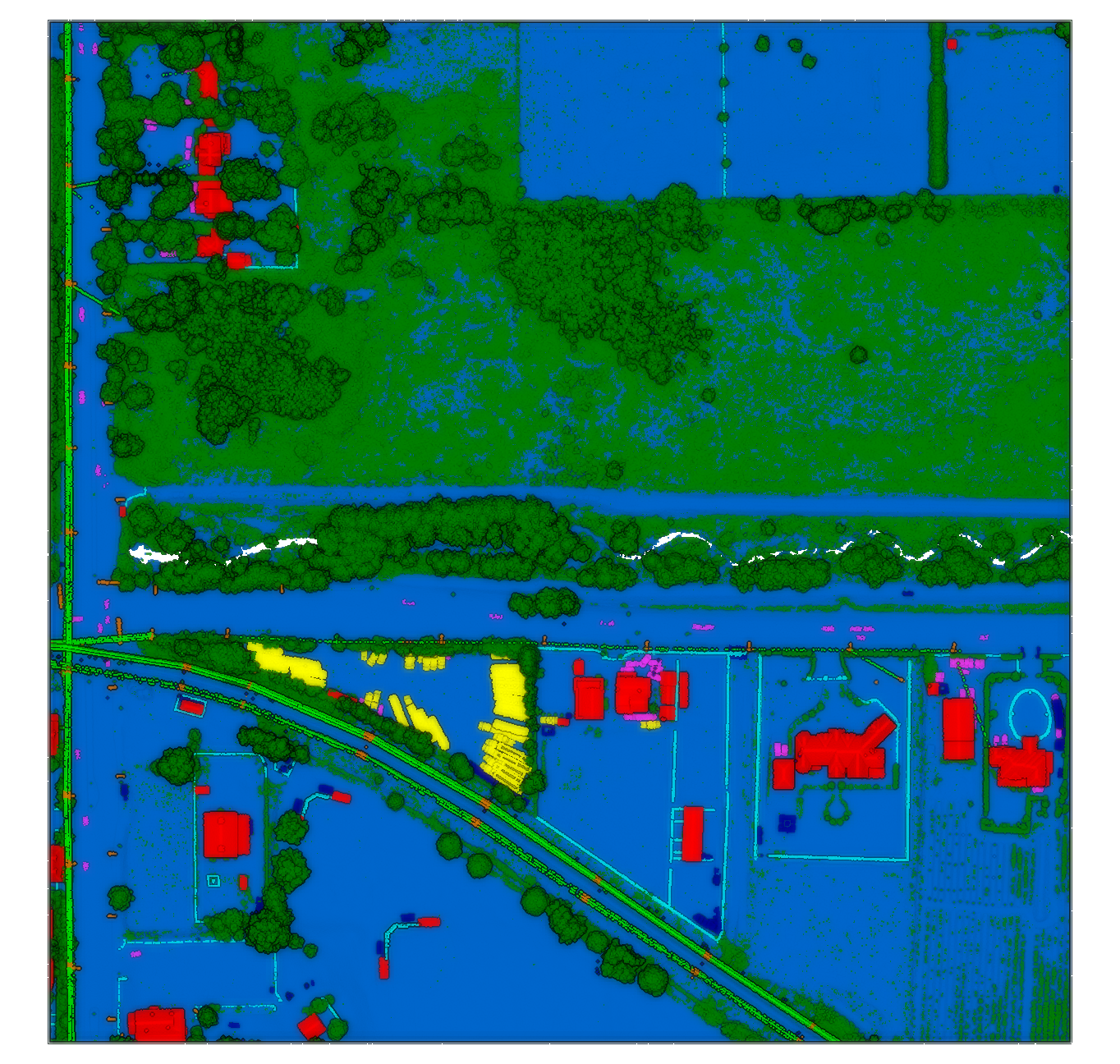}
\end{subfigure}
\begin{subfigure}{.33\textwidth}
  \centering
  \includegraphics[width=1\linewidth]{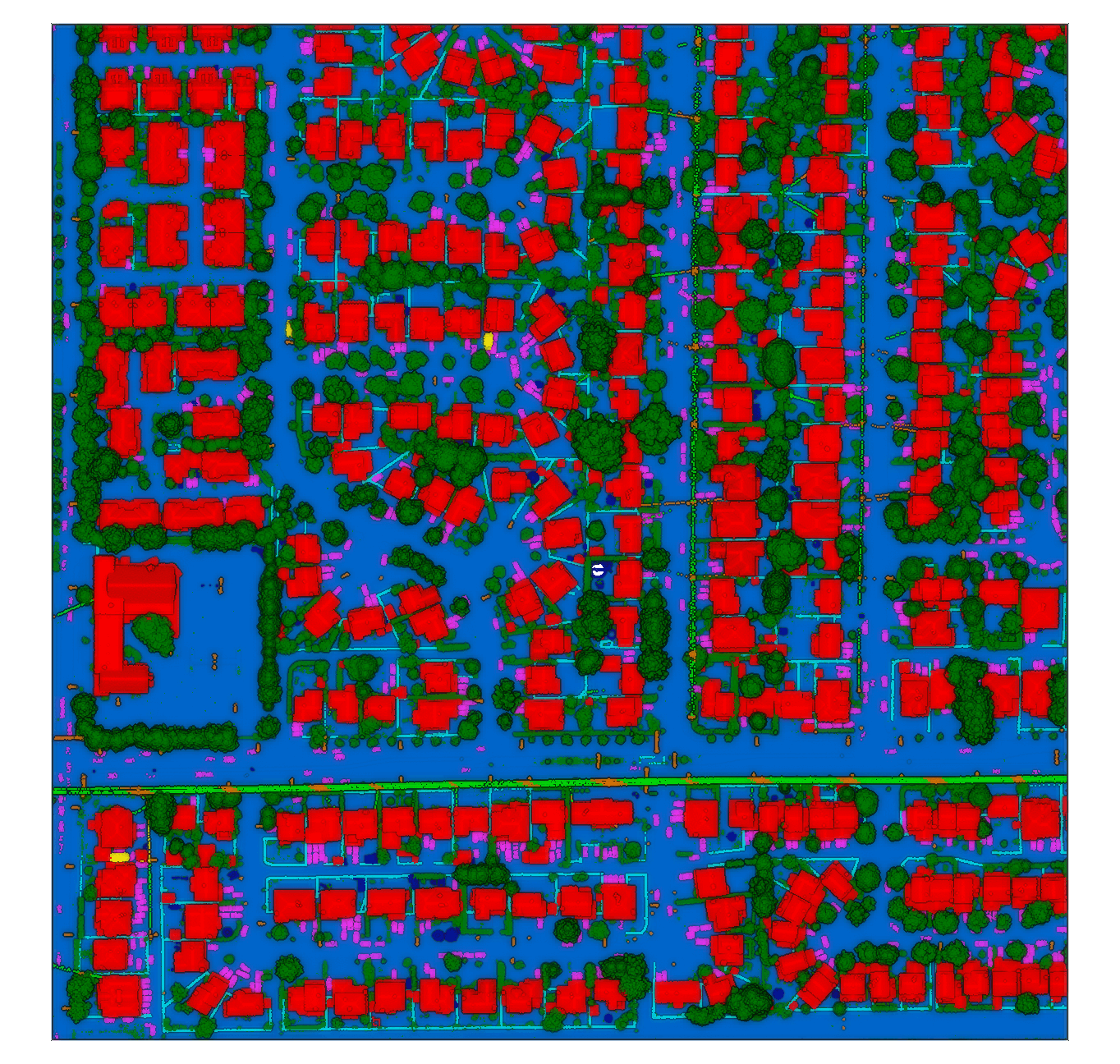}
\end{subfigure}
\begin{subfigure}{.33\textwidth}
  \centering
  \includegraphics[width=1\linewidth]{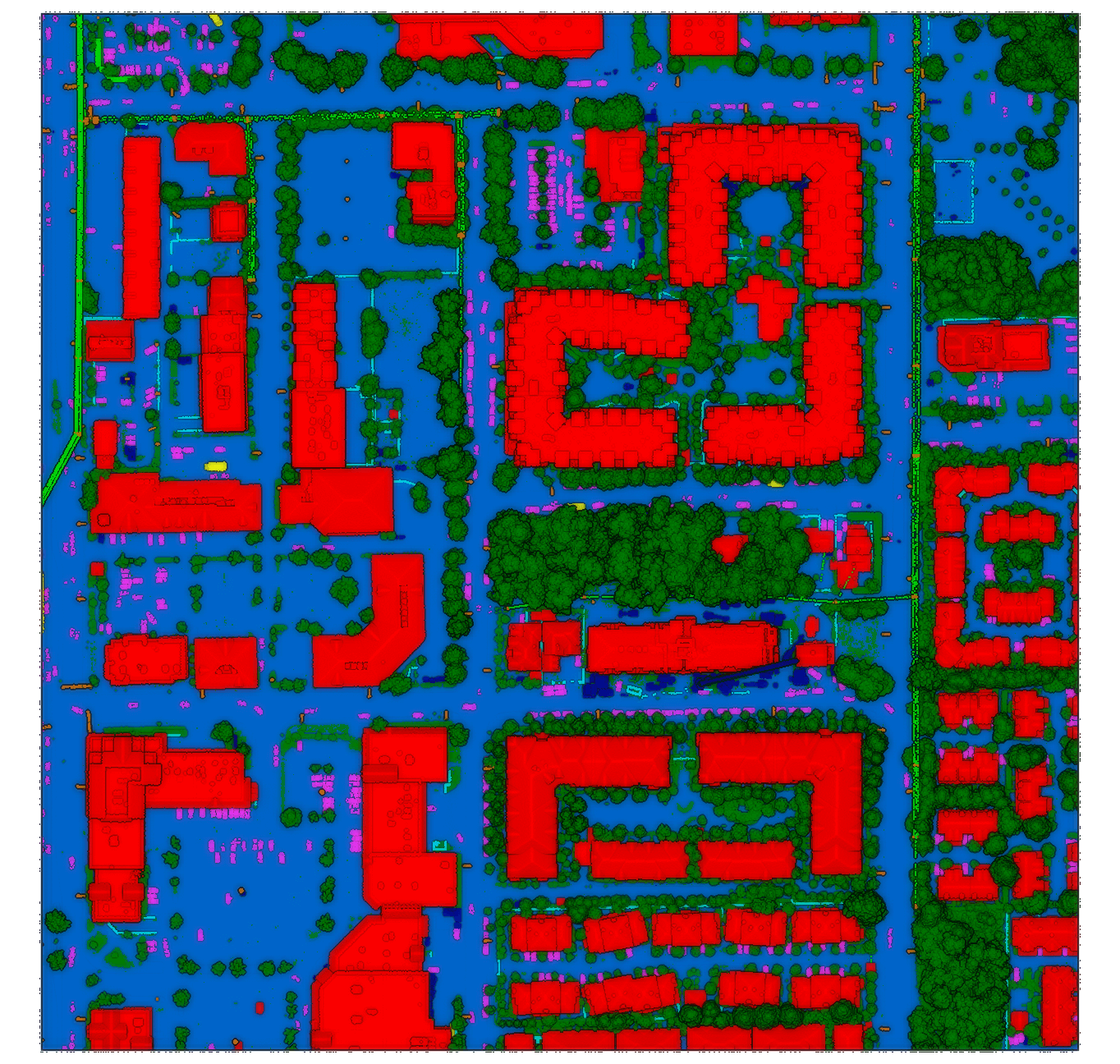}
\end{subfigure}%
\begin{subfigure}{.33\textwidth}
  \centering
  \includegraphics[width=1\linewidth]{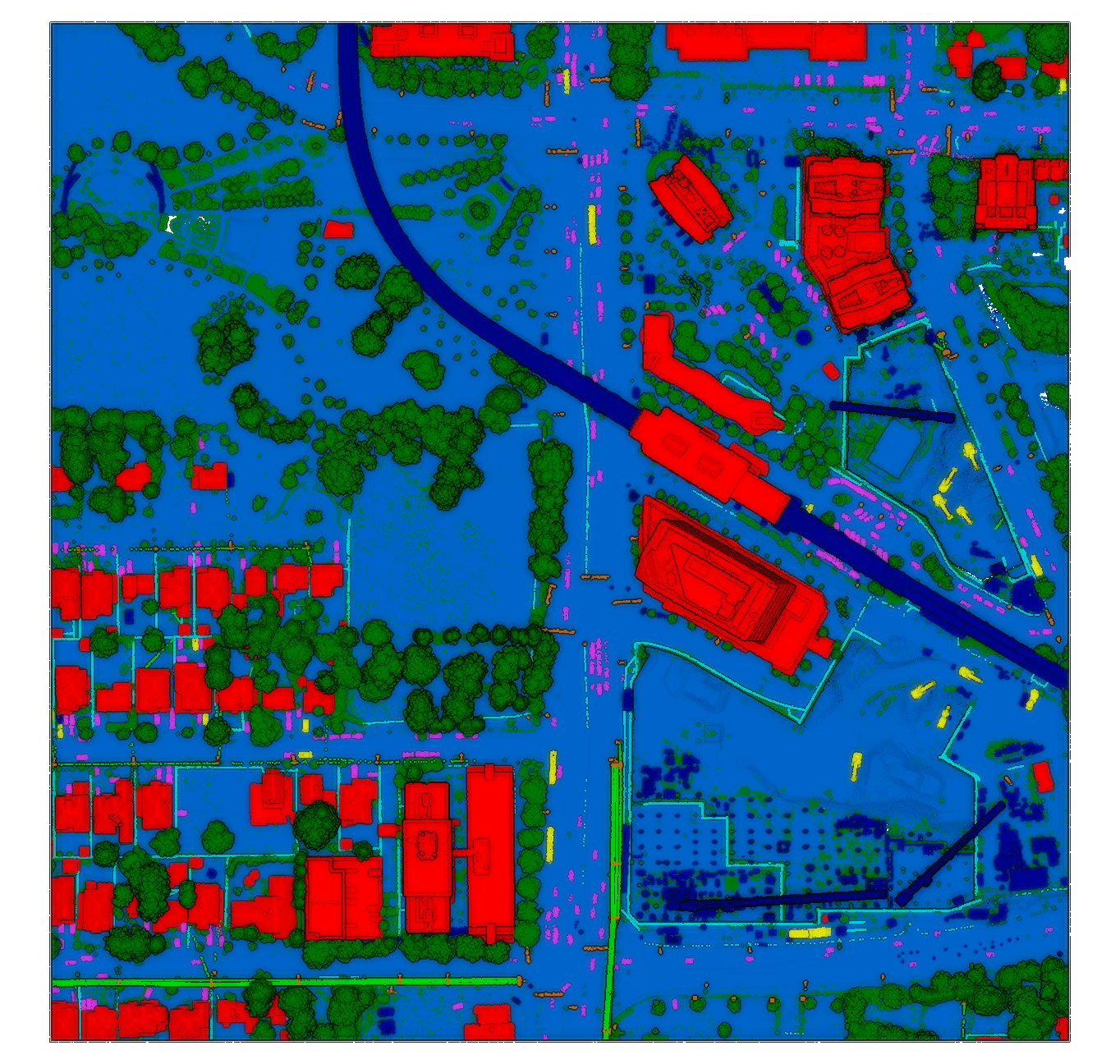}
\end{subfigure}
\begin{subfigure}{.33\textwidth}
  \centering
  \includegraphics[width=1\linewidth]{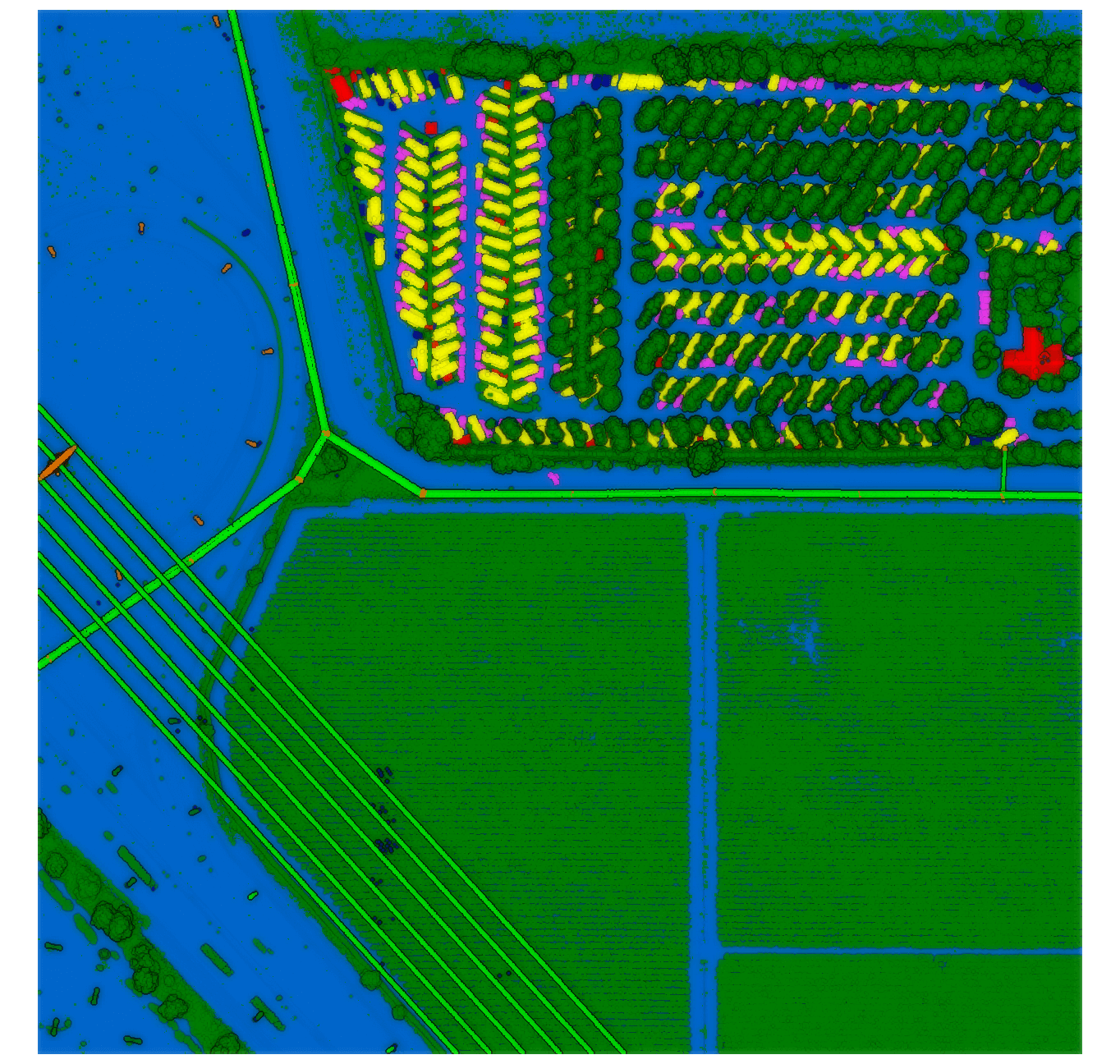}
\end{subfigure}
\caption{Example of our DALES tiles. Semantic classes are labeled by color; ground (blue), vegetation (dark green), power lines (light green), poles (orange), buildings (red), fences (light blue), trucks (yellow), cars (pink), unknown (dark blue) }
\label{fig:test-xxx}
\label{fig:2}
\end{figure*}

\subsection{Data Preparation}
We focused our initial labeling effort on a 10 $km^2$ area. The final data set consists of forty tiles, each spanning 0.5 $km^2$. On average, each tile contains 12 million points and has a resolution of 50 ppm. Unlike other data sets, our scenes do not have any overlap with neighboring tiles, so no portion of the scene is replicated in any other part of the data set, making each scene distinctly unique.

We examine all of the original tiles and cross-reference them with satellite imagery to select an appropriate mixture of scenes. Although the area is limited to a single municipality, there is a good mix of scenes and a variety of landscapes to avoid potential over-fitting. We consider four scene types; commercial, urban, rural, and suburban, which we define primarily by the type and number of buildings contained within each scene:

\begin{itemize}
\itemsep -0.18cm
\item Commercial: warehouses and office parks
\item Urban: high rise buildings, greater than four stories
\item Rural: natural objects with a few scattered buildings
\item Suburban: concentration of single-family homes
\end{itemize}
\subsection{Preprocessing}

Aerial LiDAR is subject to sparse noise within the point cloud. Noise can come from atmospheric effect, reflective surfaces, or measurement error. Despite an initial noise removal in the Terrascan software, we still found some errant noise, especially at high altitudes. Noise can have disastrous effects when performing semantic segmentation, especially with approaches that make use of voxels where a single noise point can explode the amount of memory needed for processing. De-noising is performed using a statistical outlier removal \cite{outlier} to remove sporadic points. This filter removes an average of only 11 points per tile but drastically reduces the overall bounding box in the Z direction, resulting in a reduction of 50\%.
\begin{figure}[t]
\begin{center}

    \includegraphics[width=1\linewidth]{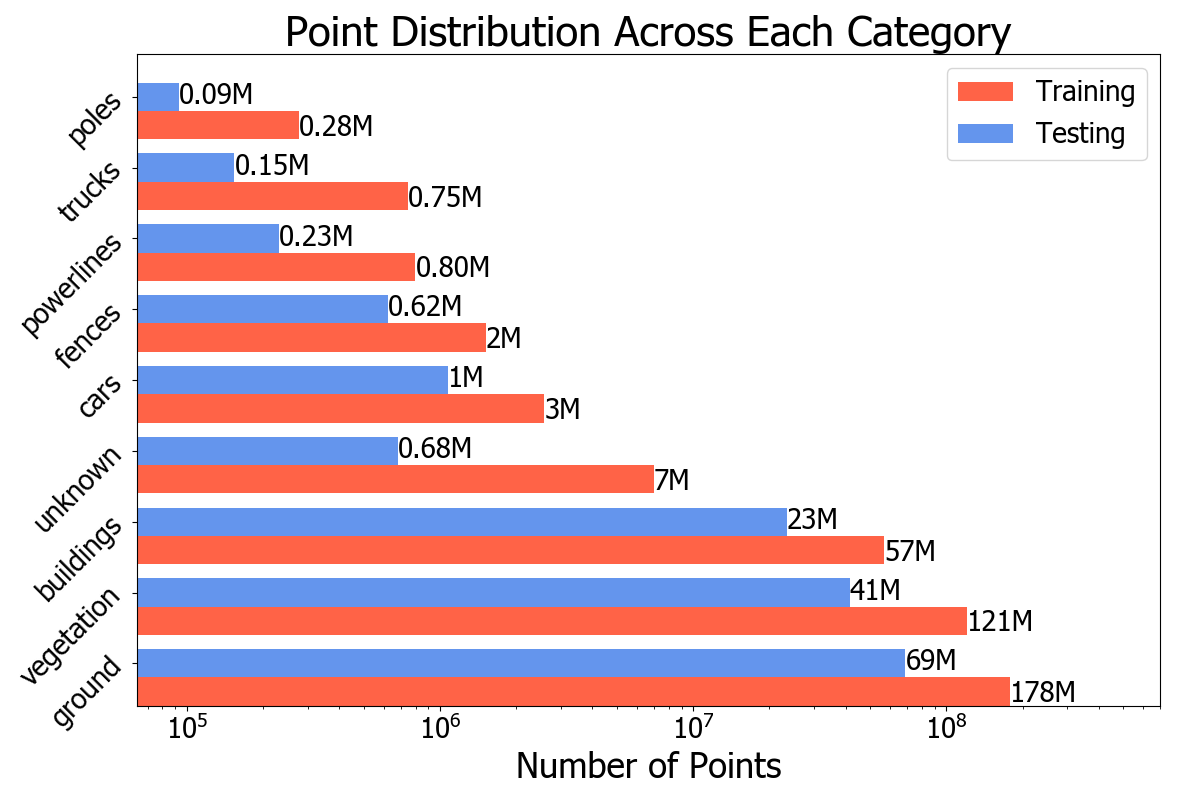}
\end{center}
   \caption{Point distribution across object categories}
\label{fig:dist}
\end{figure}
\subsection{Object Categories}
\begin{figure*}[t]
\begin{center}
    \includegraphics[width=0.99\linewidth,trim={0 10cm 0 10cm},clip]{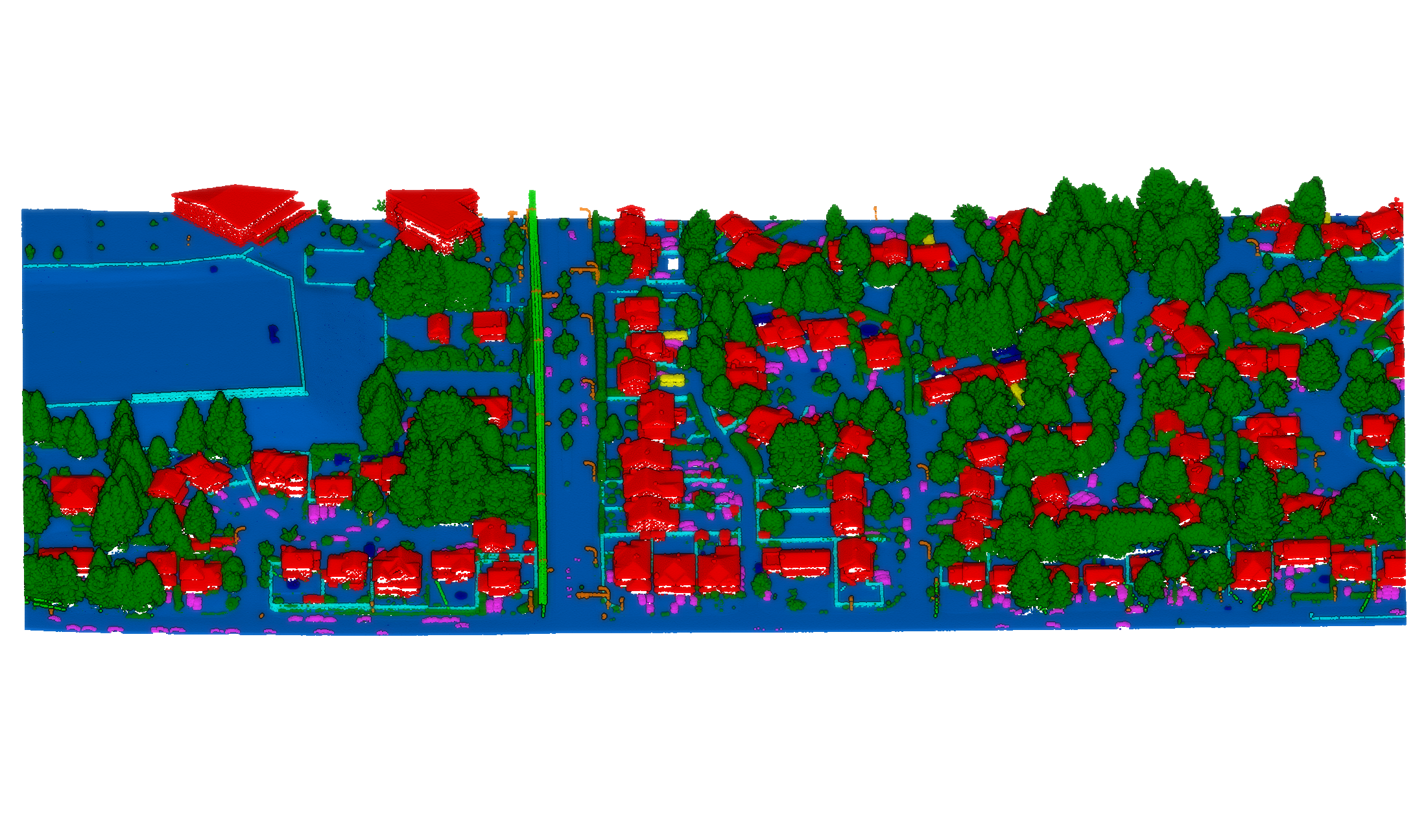}

\end{center}
   \caption{Cross section of a DALES tile. Semantic classes are labeled by color; ground (blue), vegetation (dark green), power lines (light green), poles (orange), buildings (red), fences (light blue), trucks (yellow), cars (pink), unknown (dark blue)}
\label{fig:crosssec}
\end{figure*}

We selected our object categories with a focus on appealing to a wide variety of applications, including utility monitoring, urban planning, mapping, and forestry management. We consider eight object categories, which have a wide variety of shapes and sizes. We choose our object categories to be distinct and err on the side of fewer categories to avoid object classes that are too similar to one another, such as high and low vegetation or man-made terrain versus natural terrain. The categories are as follows: buildings, cars, trucks, poles, power lines, fences, ground, and vegetation. We also include a category of unknown objects which encompass objects that were hand-labeled but did not have a critical mass of points to be effectively labeled by a deep learning algorithm. This unknown category included some classes such as substations, construction equipment, playgrounds, water, and bridges. Unknown points were labeled with a 0 and left in the scene for continuity. We do not consider unknown points in the final assessment. Figure \ref{fig:crosssec} shows a cross-section of a tile, labeled by object category. A non-exhaustive list of the types of objects included in each category is below:

\begin{itemize}
\itemsep -0.2cm
\item Ground: impervious surfaces, grass, rough terrain
\item Vegetation: trees, shrubs, hedges, bushes
\item Cars: sedans, vans, SUVs
\item Trucks: semi-trucks, box-trucks, recreational vehicles
\item Power lines: transmission and distribution lines
\item Poles: power line poles, light poles and transmission towers
\item Fences: residential fences and highway barriers
\item Buildings: residential, high-rises and warehouses
\end{itemize}

\subsection{3D Annotation}
We obtained a high accuracy DEM from the original data collection and used it to initially label all points within 0.025 meters vertical distance of the DEM as ground points. After the ground points were labeled and removed, we calculated a local neighborhood of points using K Nearest Neighbors \cite{knn}. We established a surface normal for each point using the method from \cite{pcl}. This surface normal identifies planar objects that will most likely be buildings. Using this as a starting point, manual analysis is required for refinement. About 80\% of building points are labeled using the surface normals, and the rest are hand-selected by human annotators. All objects outside of the building and ground category are exclusively hand-labeled, focusing on labeling large objects first and then iteratively selecting smaller and smaller objects. Satellite imagery, although not time synced, is also used to provide contextual information that may be useful to the annotator. Finally, we used the DEM to calculate height from ground for each point, which can also provide additional information to the human annotator. After initial annotation, the scene is confirmed by a minimum of two different annotators to ensure labeling quality and consensus between object categories.

After an initial hand labeling, we sorted the data set by object category and re-examined to check for labeling quality and consistency across labels. We used Euclidean clustering to provide a rough idea of the number of object instances. Additionally, after labeling, we decided to include extremely low vegetation ($<0.2$ meters) in the ground class to avoid confusion between object classes. The ground class encompasses both natural terrain, like grass, and man-made terrain, like asphalt.

\subsection{Final Data Format}
The final data format consists of forty tiles, 0.5 $km^2$ each, in the Airborne Laser standardized format (LAS 1.2). We provide the easting, northing, Z, and object category in their original UTM Zone 10N coordinates. We make the original coordinates available for those that may want to explore a fused data approach. For consistency with other data sets, we provide the same data as .ply files (Paris-Lille-3D data set), and .txt and .label files (Semantic 3D data set). For these files, the points follow the same order, but they contain X, Y, and Z, we set the minimum point at (0,0,0). We have randomly split the data into training and testing with roughly a 70/30 percentage split with 29 tiles for training and 11 for testing. Examples of the final labeled tiles are shown in Figure \ref{fig:2}

\section{Benchmark Statistics}
\begin{figure}[t]
\begin{center}
    \includegraphics[width=0.9\linewidth]{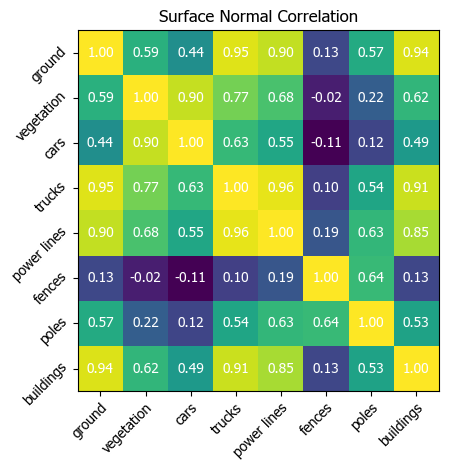}
\end{center}
   \caption{Surface normal correlation per classes}
\label{fig:correlation}
\end{figure}
We examine the data set in terms of content and complexity and make an initial assessment of the difficulty of this benchmark. Figure \ref{fig:dist} shows the approximate point distribution per class. The training and testing scenes have a similar distribution, except the unknown category, which we do not consider. As is expected, the ground class has the most points, taking up almost 50\% of the entire data set. This distribution would be the case for almost any scene, except for a dense urban city.

Vegetation is the second-largest class, with around 30\% of the entire point set. We expect this distribution because not only is vegetation a commonly occurring category, but we also decided to keep multiple returns in the data set. 

Poles have the fewest total points, making them one of the most difficult objects to classify. We performed a rough Euclidean clustering and estimated around 1100 unique pole objects in our data set; however, the nadir perspective of the sensor makes it difficult to capture many points on poles. The data set has around 340 points per pole object; these are the smallest and most challenging to detect. Additionally, every class aside from ground, vegetation, and buildings makes up less than 1\% of the data set. Object category disparity is a signature of aerial LiDAR data and is the biggest challenge when processing data.

Expanding on the problem of context and class disparity, many methods, such as PointNet and those based on it, take a set number of points from a fixed bounding box. A smaller bounding box may not have enough scene information to correctly identify the points, while a larger bounding box may not have enough points per object to detect smaller objects. This bounding box limitation is an additional challenge for methods that focus on ground-based sensors.

To get an initial assessment of the similarity between shapes in our object categories, we calculated the surface normal on a per point basis by using least-squares plane estimation in a local area. The normal orientations are flipped, utilizing the sensor location as the viewpoint. We created a histogram of these surface normals for each class and computed the correlation across all categories. Figure \ref{fig:correlation} shows the correlation heat map. Although surface normals do not totally describe the object relationships, we can see several places where strong associations occur, such as between buildings, trucks, and ground. It can be noted that because we choose several commercial areas, the truck object category includes mostly semi-trucks and box trucks, which share many geometric features with buildings. The truck category is challenging to detect due to its strong correlation with other larger object categories.
%

\begin{table*}[htbp]
\begin{center}
\begin{tabular}{lcccccccccc}
\hline & &\multicolumn{9}{|c}{IoU} \\
\hline
Method & OA &\textit{mean}& \textit{ground}& \textit{buildings}&\textit{cars}&\textit{trucks}&\textit{poles}&\textit{power lines}&\textit{fences}&\textit{veg}      \\
\hline
KPConv \cite{kpconv}& \textbf{0.978}&	\textbf{0.811}&	0.971&	\textbf{0.966}&	\textbf{0.853}&	\textbf{0.419}&	\textbf{0.750}&	\textbf{0.955}&	\textbf{0.635}&	\textbf{0.941}\\
PointNet++ \cite{pointnet++} & 0.957	&0.683	&0.941&	0.891&	0.754&	0.303&	0.400& 0.799&	0.462&	0.912\\
ConvPoint \cite{convpoint} &0.972&	0.674&	0.969&	0.963&	0.755&	0.217&	0.403&	0.867&	0.296&	0.919\\
SuperPoint \cite{superpoint} &0.955&	0.606&	0.947&	0.934&	0.629&	0.187&	0.285&	0.652&0.336&	0.879\\
PointCNN \cite{pointcnn} &0.972	&0.584&	\textbf{0.975}&	0.957&	0.406&	0.048&	0.576&	0.267&	0.526&	0.917\\
ShellNet \cite{shellnet} &0.964&	0.574&	0.960&	0.954&	0.322&	0.396&	0.200&	0.274&	0.600&	0.884\\
\hline
\end{tabular}
\caption{Overview of the selected methods on the DALES data set. We report the overall accuracy, mean IoU and per class IoU, for each category.  KPConv outperforms all other methods on our DALES data set. We also note that all methods had a large variance between object categories. }
\label{tab:results}
\end{center}
\end{table*}
\section{Evaluation}
\subsection{Metrics}

\begin{figure}[t]
\begin{center}
    \includegraphics[width=0.99\linewidth]{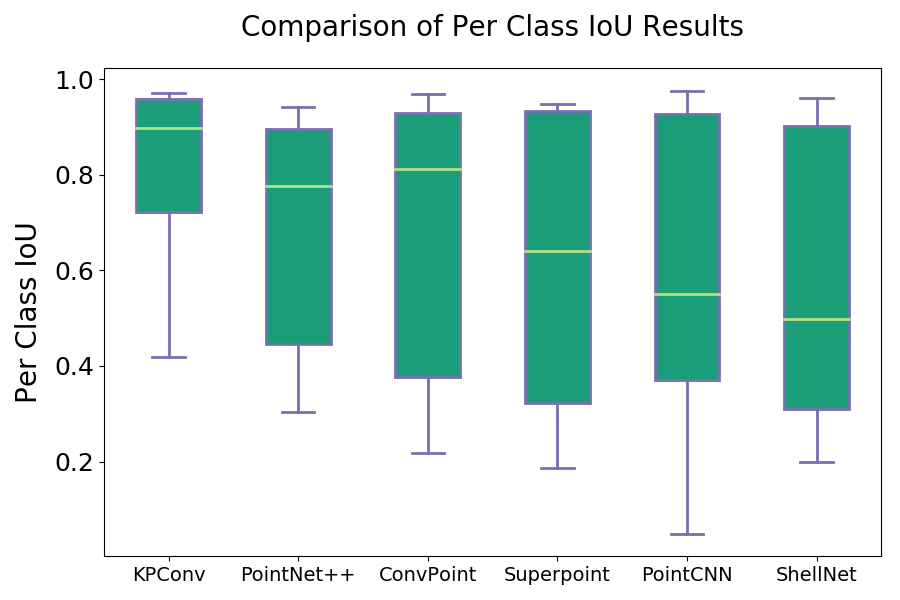}
    \end{center}
    \caption{Algorithm per class IoU performance distribution}
\label{fig:box}
\end{figure}

\begin{figure}[t]
\begin{center}
    \includegraphics[width=0.99\linewidth]{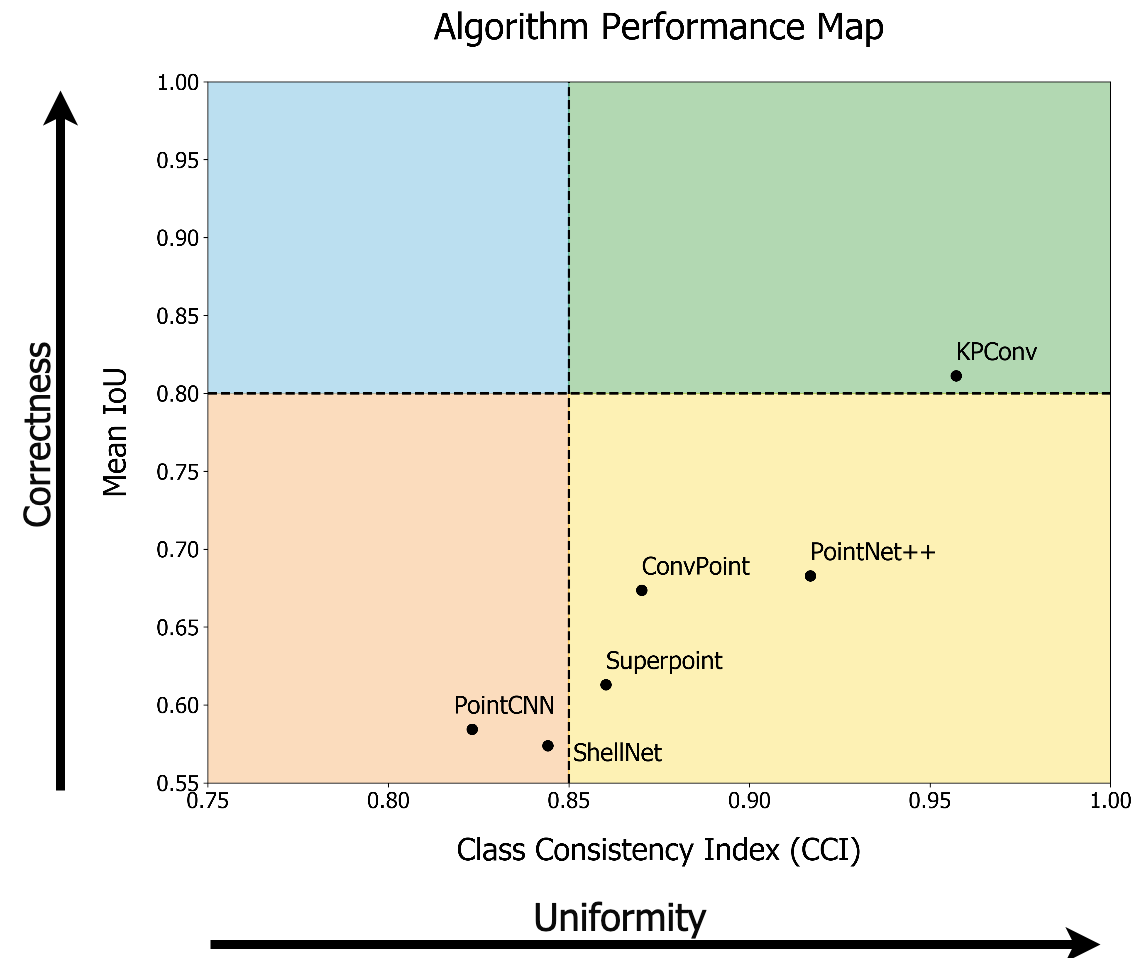}
\end{center}
   \caption{Algorithm Performance Map across six tested benchmark algorithms}
\label{fig:algo}
\end{figure}
We follow the evaluation metrics of similar MLS and TLS benchmarks and use the mean IoU as our main metric. We first define the per class IoU as: \\
\begin{equation}
    {IoU}_i = \frac{c_{ii}}{c_{ii}+ \sum\limits_{j \neq i} c_{ij} + \sum\limits_{k \neq i} c_{ki}}
\label{eq:1}
\end{equation}

The mean IoU is simply the mean across all eight categories, excluding the unknown category, of the form: \\
\begin{equation}
    \overline{IoU} = \frac{\sum\limits_{i=1}^N {IoU}_i }{N}
    \label{eq:2}
\end{equation}

We also report the overall accuracy. The measurement of overall accuracy can be deceiving when examining a data set with a significant disparity between categories. High accuracy numbers do not necessarily mean a good result across all categories, but we report it for consistency with other point cloud benchmarks. The overall accuracy can be calculated as follows:\\
\begin{equation}
    OA = \frac{\sum_{i=1}^N c_{ii} }{\sum\limits_{j=1}^N \sum\limits_{k=1}^N c_{jk}}
    \label{eq:3}
\end{equation}

We assess the results in terms of their box and whisker plots and observe the distribution between the lower and upper quartile. When evaluating the success of an algorithm, we wish to ensure that the results have both a high mean IoU and a low standard deviation across all classes. We examine the lower and upper quartiles as a measure of the robustness in the performance of a method. Finally, we establish a metric called Class Consistency Index (CCI), which we define as the complement of the in-class variance over the mean IoU, shown below: \\
\begin{equation}
    CCI = 1 - \frac{\sigma^2}{|\overline{IoU}|}
    \label{eq:4}
\end{equation}
We also examine the CCI versus the mean IoU in the form of an Algorithm Performance Map. A robust algorithm has both a high mean IoU and a high CCI, indicating that it not only has high performance but that the performance is uniform across each class.




\subsection{Algorithm Performance}
We selected six benchmark algorithms and tested their performance on our data set. We chose the algorithms based on their strong performance on similar MLS and TLS benchmarks, focusing specifically on deep learning methods that perform well in Semantic 3D and Paris-Lille-3D. We also examined the top performers in the ISPRS data set, but because these methods were not state-of-the-art, we did not select any of those methods. We selected only algorithms that have published results and available codes. The six networks we selected are PointNet++, KPConv, PointCNN, ConvPoint, ShellNet, and Superpoint Graphs. Table \ref{tab:results} shows the networks and their performance on the DALES data set.  For each network, we used the code from the original authors GitHub, and we optimized the training loss by tuning over a minimum of four runs with various parameters for each network. The best result from the multiple training runs are selected as the algorithms performance.

Overall, we find many similarities between the networks. The ground, vegetation, and building categories had strong performances over all of the networks. This strong performance is likely due to the abundance of points and examples of these categories in the data set. Alternatively, trucks, fences, and poles have a much lower IoU, which correlates to the number of points in each category.

The KPConv architecture has a notably strong performance on our data set with a mean IoU of 81.1\%, over 10\% higher than other networks. One difference between the KPConv architecture and other methods (except for Superpoint Graphs), is that KPConv did not rely on the selection of a fixed number of points within a bounding box. This method of batch selection makes it difficult to select a wide enough bounding box to adequately get scene context while also having enough points to identify small objects. In this configuration for a TLS or MLS sensor, large objects such as building walls run perpendicular to the bounding box, allowing the bounding box to contain other crucial contextual information. In our case, the large objects are building roofs that run parallel to the X and Y bounding box. In this case, a single object can take up the entire bounding box, making the points challenging to identify without additional context. We increased the size of the bounding box and also the number of points in each batch. However, this significantly increased memory and run time.

We observed consistently low performances in the truck object category. As discussed above, this category contains mostly semi-trucks and box trucks located in commercial areas. We show that the trucks have a high surface normal correlation to both the ground and building categories, both of which have significantly more training examples. This point distribution issue explains the poor performance in this object category across all methods and identifies an area for further improvement.

We also examine the box and whisker plots when evaluating network performance. From Table \ref{tab:results}, all networks perform well in our three most dominant categories: ground, vegetation, and buildings. However, performance begins to vary drastically as the number of points, and object size decreased. These differences in per class IOU results are demonstrated in Figure \ref{fig:box}. We also plot an Algorithm Performance Map shown in Figure \ref{fig:algo} and show that KPConv has the highest rating based on our mapping, both in mean IoU and CCI. We can also use the Algorithm Performance Map to make distinctions between methods with similar mean IoU performance.

We welcome the authors of the benchmarked algorithms to submit the results from their implementations. We hope that this data set is useful for the entire research community, and we look forward to additional submissions from other researchers. We deliver the full annotated data set, separated into training and testing. We also provide a website with options for downloading the data, as well as a leader board for submitting and tracking published results. The full DALES data set is available at \url{go.udayton.edu/dales3d}.
\section{Conclusion}
We presented a large scale ALS benchmark data set consisting of eight hand-labeled classes and over half a billion labeled points, spanning an area of  10 $km^2$. This data set is the most extensive publicly available aerial LiDAR data set of its kind. We evaluated the performance of six state-of-the-art algorithms on our data set. The results of the performance on this data set show that there is room for improvement in current methods, especially in their ability to evaluate semantic segmentation in classes of different physical sizes and number of points. We hope that this benchmark can be a resource for the research community and help advance the field of deep learning within aerial LiDAR. For future work, we will continue this labeling effort to include more classes and eventually encompass the entire 330 $km^2$ area and present this data as a challenge to the earth vision community. 

\section{Acknowledgement}
The data set presented in this paper contains Information licensed under the Open Government License – City of Surrey.

{\small
\bibliographystyle{ieee_fullname}
\bibliography{egbib}
}

\end{document}